# Facial Action Unit Detection on ICU Data for Pain Assessment

**Subhash Nerella[a], Azra Bihorac[b], Patrick Tighe[c], Parisa Rashidi[a]**

a- Biomedical Engineering, University of Florida

b- Division of Nephrology, University of Florida

c- Department of Anesthesiology, University of Florida

## ABSTRACT

Current day pain assessment methods rely on patient self-report or by an observer like the Intensive Care Unit (ICU) nurses. Patient self-report is subjective to the individual and suffers due to poor recall. Pain assessment by manual observation is limited by the number of administrations per day and staff workload. Previous studies showed the feasibility of automatic pain assessment by detecting Facial Action Units (AUs). Pain is observed to be associated with certain facial action units (AUs). This method of pain assessment can overcome the pitfalls of present-day pain assessment techniques. All the previous studies are limited to controlled environment data. In this study, we evaluated the performance of OpenFace an open-source facial behavior analysis tool, and AU R-CNN on the real-world ICU data. Presence of assisted breathing devices, variable lighting of ICUs, patient orientation with respect to camera significantly affected the performance of the models, although these showed the state-of-the-art results in facial behavior analysis tasks. In this study, we show the need for automated pain assessment system which is trained on real-world ICU data for clinically acceptable pain assessment system.

## 1 Introduction

Pain is a defensive mechanism of body raising alarm to take preventive action to impending damage to the body. Pain assessment is necessary to prevent disease progression and faster recovery of patients. Pain experience and expression are subjective to the individual, which makes it difficult to diagnose pain. There is no well-defined objective scale for identifying the pain, doctors depend on self-reported pain by individual subjects or manual observation by staff. Self-reported pain is the gold standard for pain assessment. Patients in the Intensive care unit and post-surgery cannot self-report pain due to various reasons like being on ventilators, the influence of sedatives, and being unconscious. Untreated pain could potentially result in multiple complications and seriously impact patient chances of well-being. Incomplete or no assessment of pain in ICU has been associated with death [1]. Pain has an impact on the economy through the loss of productive employees and the cost of care for chronic pain[2]. The cost of chronic pain to society is more than cancer, HIV, and heart disease combined[2]. The financial impact of pain on the United States economy was estimated at $100 billion a year[3].

Face plays a prominent role in nonverbal communication [4, 5]. Various applications in automotive industries, education, surveillance, and entertainment uses facial behavior to facilitate human-computer interaction[6-8]. These applications are successful in capturing facial behavior suggest feasibility for automated pain detection based on facial cues to detect pain. Behavioral scientists devised a reliable approach based on facial indicators of pain called action units (AUs). Facial Action Coding System (FACS) is a facial anatomy-based action coding system, which comprises of different facial action units with intensity [9]. Any facial expression can be represented as a combination of these facial action Units. Prkachin and Solomon[10] identified certain facial action units are associated with pain developed a metric Prkachin Solomon Pain Index (PSPI) score based on the facial action coding system. A major caveat of this approach is that it demands trained personnel's supervision of patients. It must be performed manually which makes it both time consuming and costly in turn making it clinically unviable option.

An automated real-time pain detection system would potentially make it clinically feasible and enhance patient care. Identification of action units (AUs) is an essential step in the development of an automatic pain assessment system. The major factor hindering the development of an automated pain detection system until recent years is the lack of well-annotated representative data. Researchers at McMaster University and the University of Northern British Columbia (UNBC) captured videos of patient's faces suffering from shoulder pain[11]. Each video frame was fully action unit (AU) coded by certified facial action unit coding system (FACS) coders[9]. New datasets like BP4D-Spontaneous[12] and BP4D+[13] are available with pain data recorded from healthy volunteers but the pain is stimulated by cold presser task.

All the datasets available are limited to a controlled environment. To our knowledge, there is no literature available showing the performance of the models developed based on these controlled environment datasets work on real-time ICU data. Our lab Intelligent health systems collected video data from ICUs at the University of Florida. In this study, we evaluated the performance of the open-source facial action unit's detection on ICU data. We used OpenFace [14] an open-source facial behavior analysis tool to detect behavior, and AU R-CNN[15] to determine the facial action units present in the image frames extracted from the ICU videos. OpenFace although it showed state of the art results on facial behavior analysis, failed to generalize well on the ICU data. We further trained a model based on AU R-CNN architecture on the ICU dataset showing better results. In this study, we show the requirement of a system trained on the ICU data to detect facial action units for the objective of pain assessment in a clinical setting.

## Methods

All the data used in this study were collected at surgical ICUs at the University of Florida Health Hospital, Gainesville, Florida. The study was reviewed and approved by the University of Florida Institutional Review Board. All the methods were performed with written informed consent obtained from patients before enrollment in the study, following guidelines, and regulation of IRB. All the images of patients analyzed for this paper are from patients who have given written consent to publish the data for research purposes.

| Action Unit | Description | Our's | OpenFace | AU-RCNN |
|---|---|---|---|---|
| AU 01 | Inner Brow Raiser |  | ✓ | ✓ |
| AU 02 | Outer Brow Raiser |  | ✓ | ✓ |
| AU 04 | Brow Lowerer | ✓ | ✓ | ✓ |
| AU 05 | Upper Lid Raiser |  | ✓ |  |
| AU 06 | Cheek Raiser | ✓ | ✓ | ✓ |
| AU 07 | Lid Tightener | ✓ | ✓ | ✓ |
| AU 09 | Nose Wrinkler | ✓ | ✓ |  |
| AU 10 | Upper Lip Raiser | ✓ | ✓ | ✓ |
| AU 12 | Lip Corner Puller | ✓ | ✓ | ✓ |
| AU 14 | Dimpler |  | ✓ | ✓ |
| AU 15 | Lip Corner Depressor |  | ✓ | ✓ |
| AU 17 | Chin Raiser |  | ✓ | ✓ |
| AU 20 | Lip Stretcher | ✓ | ✓ |  |
| AU 23 | Lip Tightener |  | ✓ | ✓ |
| AU 24 | Lip Pressor | ✓ |  | ✓ |
| AU 25 | Lips Part | ✓ | ✓ |  |
| AU 26 | Jaw Drop | ✓ | ✓ |  |
| AU 27 | Mouth Stretch |  | ✓ |  |
| AU 28 | Lip Suck |  | ✓ |  |
| AU 43 | Eyes Closed | ✓ |  |  |
| AU 45 | Blink |  | ✓ |  |

Table1 List of action units specific to different approaches. Green check mark represents the AU is applicable. Only the AUs that are common between the annotated ICU data and the approach are evaluated.

To collect video data from ICU, a standalone system with two cameras mounted was used, these cameras captured ICU rooms with a patient. Fifteen-minute videos were extracted which are further processed to obtain individual frames. In total 900 frames were extracted from each video. For this study we were interested in the patient's facial behavior, so we extracted patient faces from the video. All the video frames were processed using Facenet [16] to detect and crop patient faces from the input video frames. It can be seen how real-world ICU data can be from controlled environment data, like the effects of sedatives, lighting in a room, presence of ventilators, and face orientation with respect to the camera.

Three annotators were recruited and trained on the Facial Action Coding System (FACS) coding system. Annotators were evaluated on sample images before they performed annotation on ICU video frames. Table1 shows AUs used by previous approaches and our dataset. All the annotations were performed individually on the images assigned. Mode of the multiple annotator's labels is considered as ground truth label i.e. whether a particular action unit is present or absent in each image. We used Fleiss' kappa[17] score to assess the annotator's agreement on the images. Only images that are annotated by three annotators are considered and further images where at least two out of three annotators had agreed are used for evaluating OpenFace and AU Nets. The mean Fleiss' kappa score of 0.48 was obtained based on the images annotated. In table 2, facial action units are listed with their corresponding descriptions. We have used video data of 13 patients from ICU in this paper. We extracted 151,947 image frames from all the patients out of which we considered 55,085 images which are marked as clear by all the annotators and with the good agreement are selected. All the 55,085 images were evaluated by Openface. Although we collected multiple AUs, in this work we only limited our analysis to AUs 25, 26 and 43.

## Results

All the ICU videos are processed to obtain the patient's faces from every frame. These extracted faces are provided as an input to 1. OpenFace [14] an open-source tool available for facial behavior analysis using computer vision and Machine learning algorithms. 2. AUR-CNN a convolution network approach for recognition of facial action units. Although the objective of these approaches is the same (facial action unit detection) they have a different approach to the problem. In the pipeline of identifying facial action units, OpenFace detects facial landmarks which are further used in facial alignment, feature fusion, and ultimately facial action unit detection. Most approaches used facial landmarks to align the images to further detect facial AUs. Table1 shows the AUs OpenFace and AUR-CNN are capable of capturing. We evaluated the performance of these models on the AUs 25,26 and 43.

AU R-CNN[15] proposed by Chen et al uses Faster R-CNN[18] based approach to predict the presence of AUs in an image. This approach achieved state of the art results on BP4D dataset[12]. The network architecture proposed consists of two modules, feature extraction module, and the head module. Feature extraction module takes an individual face image as input and outputs the features. A region of Interest is localized to an AU is extracted from the features and provided as input the head module. The region of interest is extracted based on facial landmarks. Dlib[19] library provides 68 landmarks on the face. The coordinates of the landmarks are used to extract relevant region of interest for each Action unit on the face. Some of the AUs share a common region of interest so in total nine regions of interest are extracted the corresponding occurrence of facial action units. The Head module is the classifier that predicts the presence of the AU. The feature extractor module consists of (res1, res2, res3, res4) blocks of resnet101 architecture. The head module comprises (res5, average pool, and fc layers). The last fully connected layer is modified to have a size equal to the number of AUs being predicted. The network uses sigmoid cross entropy loss as multiple AUs can be present in given region of interest. We trained the model pretrained on BP4D dataset and further trained on our ICU datasets and evaluated the performance of the trained model.

| Action Unit | accuracy | f1score | support |
|---|---|---|---|
| AU25 | 0.51 | 0.49 | 32587 |
| AU26 | 0.58 | 0.34 | 23827 |

Table2 shows OpenFace performance on entire ICU data reported for each action unit. OpenFace performance is evaluated against ground truth annotation.

| Action Unit | Accuracy | F1-score | Support |
|:---:|:---:|:---:|:---:|
| AU25 | 0.69 | 0.79 | 3396 |
| AU26 | 0.65 | 0.75 | 2874 |
| AU43 | 0.60 | 0.63 | 3752 |

Table3 shows Trained AU-RCNN performance on ICU test set reported for each action unit. AU-RCNN performance is evaluated against ground truth annotation.

| Action Unit | accuracy | f1score | support |
|:---:|:---:|:---:|:---:|
| AU25 | 0.39 | 0.12 | 3396 |
| AU26 | 0.41 | 0.1 | 2874 |

Table4 shows OpenFace performance on ICU test set reported for each action unit. OpenFace performance is evaluated against ground truth annotation.

## Discussion

OpenFace can detect facial landmarks, head pose estimation, eye-gaze detection, and facial action unit recognition. In this study, we are only interested in facial action unit detection as these units play a key role in pain detection. We reported the performance of OpenFace on ICU data in Table2 with respect to action units for all images with a good annotation agreement. OpenFace does not predict AU43 (Eyes closed). Overall OpenFace struggled to capture AU25 (Lips Part) and AU26(Jaw Drop) shown in Table 3. In the pipeline of OpenFace prediction detecting facial landmarks is one of the initial steps. Accurate detection of these facial landmarks plays a crucial role in the performance of the tool to predict facial action units. OpenFace uses a combination of predicted landmarks and Histograms of the gradient to predict the AUs. Inaccurate prediction of these landmarks results in poor performance of the model in prediction of action unit occurrence. The presence of medical devices and inconsistent face orientation affected the prediction of landmarks. Openface is trained on the datasets in which data is collected from the controlled lab environment. Controlled environment images have good light intensity, subject face orientation with respect to the camera, and no occlusion to the face. Realtime ICU environment cannot ensure the ideal conditions of a controlled environment. All the above factors effected the performance of OpenFace on the ICU data.

Table3 shows the performance of AU R-CNN on the ICU data. We started with a pre-trained model on the BP4D dataset to further train on the ICU image data. ICU dataset is randomly split into three partitions train, validation and test ensuring no patients data is common between partitions. The model is trained for 20 epochs with sigmoid cross-entropy loss. The trained model is evaluated on the test partition and the corresponding result is shown in Table3. The model performed well in the case of AUs 25,26,43. This trained model showed better performance than Openface for AU25 and AU26 on the same test set shown in Table4. The reason for better prediction of these AUs is due to their strong presence in the dataset in terms of number of images, which aided the network during training. Patients in ICU are more likely to be under medication which effected the individuals being less responsive in facial expression can be the reason for the low presence of other action units. The F1-score of AU 25, 26 and 43 exceeded the accuracy which shows that model overestimated the presence of these AUs. This trained model showed how training the models can improve the prediction of Acton units in ICU images. A limitation of this approach is Dlib failed to capture facial landmarks of some of the ICU images due to the low resolution of ICU images. As ICU is a restricted environment, cameras cannot be placed close to the patient's face which resulted in poor resolution of images. In the future we will switch to a better landmark prediction model and consider other AUs to our evaluation.

Prkachin Solomon Pain Index (PSPI score) [10] takes action units AU4, AU6, AU7, AU9, AU10, and AU43 into account, presence of the AUs also their corresponding intensity on a scale of (0-5) to yield a 16 point pain scale. To assess pain in terms of the PSPI score requires an automated AU detection tool. AU 43 (Eyes closed) is not captured by both OpenFace and AU nets but it is easy to detect. AU43 is sensitive to pain but not specific to it. PSPI score is a standard for measuring pain but the non-zero PSPI score does not necessarily mean the person is in pain because Several action units share presence in different facial expressions like pain and happiness (AU 6), fear (AU 4), disgust (AU 10/9). A secondary ground truth like a nurse's diary of pain score can help achieve a better model. In the future, we intend to collect more data from ICU to develop a better model for the prediction of Action Units that can provide more information about the pain in the ICU environment. Our dataset so far is limited in the presence of AUs other than AU 25,26 and 43.

**Conclusion**

Automatic pain assessment is beneficial to health care in terms of both cost and efficiency. Pain assessment through facial expression requires the detection of facial action units. In this paper, we evaluated the performance of OpenFace, and AU R-CNN models on actual ICU data in predicting the presence of facial action units. The ICU data being different from the data OpenFace trained on resulted in poor performance of the model. The model struggled to identify the AUs although they achieved good performance on the controlled environment datasets. A lot of factors like the presence of medical assist devices presence on patient's faces, lighting in the ICU, image quality and patient face orientation resulted in poor performance of these models. Based on the performance of OpenFace and AU nets on real ICU data we conclude that to achieve automatic pain detection on real ICU data requires the model to be trained on ICU data to achieve the end goal of automated pain assessment.